\documentclass[letterpaper, 10 pt, conference]{ieeeconf}  % Comment this line out if you need a4paper

\IEEEoverridecommandlockouts                              % This command is only needed if 
\usepackage{graphics} % for pdf, bitmapped graphics files
\usepackage{epsfig} % for postscript graphics files
\usepackage{mathptmx} % assumes new font selection scheme installed
\usepackage{times} % assumes new font selection scheme installed
\usepackage{amsmath} % assumes amsmath package installed
\usepackage{amssymb}  % assumes amsmath package installed
\usepackage{ulem}
\usepackage{multirow}
\usepackage{booktabs}
\usepackage{tabularx}
\usepackage{comment}
\usepackage{tablefootnote}

\usepackage{xcolor}
\usepackage{algorithm}
\usepackage{algpseudocode}
\usepackage[colorlinks=False]{hyperref}

\usepackage[symbol]{footmisc}

\newcommand{\jiahui}[1]{{\color{black} #1}}
\newcommand{\jhs}[1]{{\color{black} #1}}%{{\color{cyan} #1}}
\newcommand{\edward}[1]{{\color{black} #1}}%{{\color[rgb]{0.7,0.2,0.7}#1}}

\def\ie{i.e.}

\newcolumntype{C}[1]{>{\centering\arraybackslash}p{#1}}

\title{\LARGE \bf
SDF-Pack: Towards Compact Bin Packing with\\ Signed-Distance-Field Minimization
}

\author{Jia-Hui Pan, Ka-Hei Hui, Xiaojie Gao, Shize Zhu, Yun-Hui Liu, Pheng-Ann Heng, and Chi-Wing Fu 
\thanks{\jiahui{This work is supported in part by the Shenzhen Portion of Shenzhen-Hong Kong Science and Technology Innovation Cooperation Zone under HZQB-KCZYB-20200089 and the InnoHK of the Government of the Hong Kong Special Administrative Region via the Hong Kong Centre for Logistics Robotics.}
J.-H. Pan, K.-H. Hui, X. Gao, P.-A. Heng, and C.-W. Fu are with the Department of Computer Science and Engineering, The Chinese University of Hong Kong. S. Zhu is with the Hong Kong Centre for Logistics Robotics. Y.-H. Liu is with the Department of Mechanical and Automation Engineering, The Chinese University of Hong Kong.}
}

\begin{document}

\maketitle
\thispagestyle{empty}
\pagestyle{empty}

%%%%%%%%%%%%%%%%%%%%%%%%%%%%%%%%%%%%%%%%%%%%%%%%%%%%%%%%%%%%%%%%%%%%%%%%%%%%%%%%
\begin{abstract}
Robotic bin packing is very challenging, especially when considering practical needs such as object variety and packing compactness.
This paper presents SDF-Pack, a new approach based on signed distance field (SDF) to model the geometric condition of objects in a container and compute the object placement locations and packing orders for achieving a more compact bin packing.
Our method adopts a truncated SDF representation to localize the computation, and based on it, we formulate the SDF-minimization heuristic to find optimized placements to compactly pack objects with the existing ones.
To further improve space utilization, if the packing sequence is controllable, our method can suggest which object to be packed next.
%.
Experimental results on a large variety of everyday objects show that our method can consistently achieve higher packing compactness over 1,000 packing cases, enabling us to pack more objects into the container, compared with the existing heuristics under various packing settings.
The code is publicly available at: \href{https://github.com/kwpoon/SDF-Pack}{https://github.com/kwpoon/SDF-Pack}.
%{\it We will release the code of SDF-Pack after the publication of this work.}
%
\end{abstract}

%%%%%%%%%%%%%%%%%%%%%%%%%%%%%%%%%%%%%%%%%%%%%%%%%%%%%%%%%%%%%%%%%%%%%%%%%%%%%%%%
\section{INTRODUCTION}
Robotic bin packing is an important logistics task, aiming at leveraging robot arms to help automatically pack objects in a container.
Given a sequence of objects of arbitrary sizes and shapes, a bin-packing algorithm should suggest the optimized placement location of each object, such that the objects can be packed compactly in the container.
If the sequence is controllable, the algorithm may further suggest which object to be packed next in each iteration.

At present, widely-used basic algorithms focus mainly on regularly-shaped objects, e.g., boxes.
Packing irregularly-shaped objects still heavily relies on manual efforts, since several challenges remain.
First, it is difficult to effectively model the container geometry, especially when considering varying objects of irregular shapes.
Second, it is difficult to exploit the geometry of irregular objects, and also the container, to search for a compact bin-packing solution that maximizes space utilization.
Last, the search should be sufficiently fast to avoid long delays in the robotic actions.

For the above practical concerns, packing heuristics or strategies~\cite{baker1980orthogonal,karabulut2004hybrid,wang2019stable,wang2010two} have long been a favorite for robotic bin packing, especially for handling general irregular objects, since these approaches are fast to compute.
Basically, they assume a predefined packing order and suggest a compact packing location for each object according to a designed objective function.
Yet, existing heuristics cannot sufficiently account for the compactness between the new object to be packed and the existing ones in the container, so the first two challenges cannot be well addressed.
For example, \jiahui{the} deepest bottom-left~\cite{baker1980orthogonal,karabulut2004hybrid} places \jiahui{the object} as deep and bottom-left as possible, without modeling the object geometry;
heightmap minimization~\cite{wang2019stable} aims to reduce the overall heightmap value, without considering the compactness between the new object and the existing ones at the same height level;
while maximum touching area~\cite{wang2010two} encourages direct object contacts, without attending to non-touching but very close packing. 
Hence, they have limited capability to handle irregular objects for better space utilization.

\begin{figure}[t]
\begin{center} \includegraphics[height=0.54\linewidth, width=\linewidth]{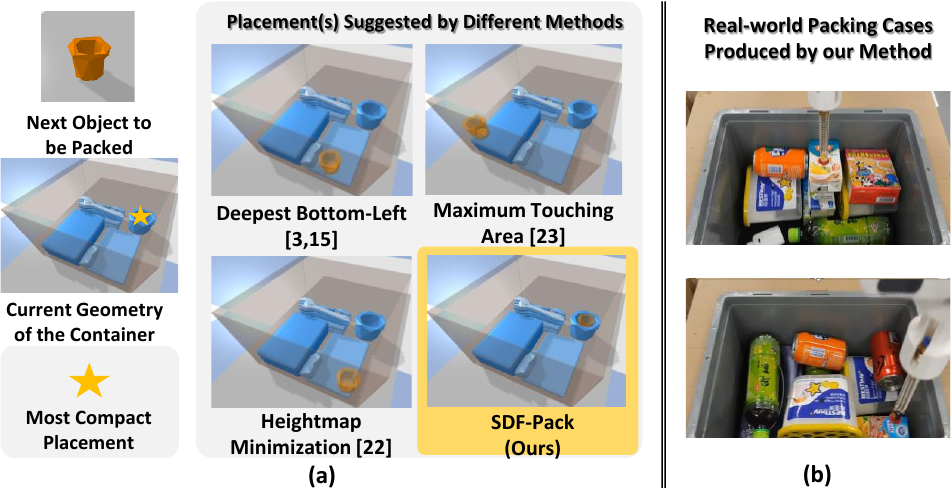}
\end{center}
\vspace{-10pt}
\caption{
\jhs{(a)} Comparing with the existing heuristics, our method is able to suggest the most compact object placement location by using our formulated SDF-minimization heuristic to efficiently assess the spatial compactness in the container and encourage closely-packed object placement locations. \jhs{(b) Two real-world packing cases produced by our method.}}
\vspace{-12pt}
\label{fig:teaser}
\end{figure}

This paper presents SDF-Pack, a new approach to address the challenges for compact packing of general objects in a container; 
see the illustration in Figure~\ref{fig:teaser}. 
We adopt the signed distance field (SDF)~\cite{gibson1998using,fedkiw2002level} to model the container's geometry, assess the spatial compactness in the container, and formulate the SDF-minimization heuristic to locate compact object placements.
In particular, we first construct an SDF \jhs{using a scanned top-down heightmap of the container}, a volume field that records the shortest distance from each 3D location in the container to the nearest object/container surface.
Using the constructed SDF, we can quickly identify locations with minimum non-negative SDF values and find collision-free locations that compactly pack the next object with the existing ones.
Further, by comparing the SDF heuristic values of different candidate objects, our proposed algorithm may suggest which object to be packed next, if the sequence is controllable.
Overall, the contributions of this work are summarized below:
\begin{itemize}
\item[(i)]
We introduce SDF-Pack, a new approach based on the truncated signed distance field to model the geometry of the container and packed objects.
Importantly, we formulate the SDF-minimization heuristic to find the most compact placement for a given object to improve space utilization and packing compactness.
\item[(ii)]
For controllable packing sequence, SDF-Pack can suggest the next object to be packed to further improve the overall object packing compactness.
\jhs{\item[(iii)]
To speed up our framework, we further develop a GPU implementation to build the SDF and a local update scheme to avoid redundant computation.
}
\end{itemize}

We compare SDF-Pack with the state-of-the-art heuristics~\cite{baker1980orthogonal,karabulut2004hybrid,wang2019stable,wang2010two} on general irregular objects from the YCB dataset~\cite{ycb} and the Rutgers APC RGB-D dataset~\cite{apc}.
Experimental results show that SDF-Pack is able to {\it{consistently achieve higher packing compactness}} over 1,000 packing cases under various packing settings.

%%%%%%%%%%%%%%%%%%%%%%%%%%%%%%%%%%%%%%%%%%%%%%%%%%%%%%%%%%%%%%%%%%%%%%%%%%%%%%%%
\section{RELATED WORK}
3D bin packing (3D-BPP) is an NP-hard problem, generally requiring nondeterministic polynomial time to solve. 
It involves two sub-problems:
(i) finding the object packing order and 
(ii) finding the placement location of each object.

Many methods are proposed to optimize both (i) and (ii) simultaneously.
Since the problem is inherently NP-hard, computational time remains an issue.
Some exact methods, e.g., branch-and-bound~\cite{martello2000three}, try to reduce the search space for the optimal solution.
Instead of exhaustively searching the whole solution space, meta-heuristic methods such as the genetic algorithm~\cite{ramos2016container, karabulut2004hybrid, kang2012hybrid}, tabu search~\cite{crainic2008extreme}, simulated annealing~\cite{liu2015hape3d}, and integer linear programming~\cite{lamas2022voxel, jiang2012learning} search for sub-optimal solutions. Yet, these methods require trying multiple packing sequences and thus are still computationally expensive.
Very recently, reinforcement learning is introduced for 3D-BPP \cite{hu2020tap, duan2018multi, zhang2021attend2pack, huang2022planning, zhao2022learning}\jiahui{, and in particular, \cite{huang2022planning, zhao2022learning} attempt to pack non-regular objects. Reinforcement learning methods help to optimize the multi-step packing results.}
However, they require an extra \jiahui{training stage to create a network model and train on sampled packing sequences}, \jiahui{e.g.,~\cite{huang2022planning} took 16 hours of training}.

On the other hand, some other works simplify the problem by assuming a user-defined packing sequence and focus mainly on (ii).
They propose data structures such as Markov decision tree~\cite{yang2021packerbot,zhao2021online} and packing configuration tree~\cite{zhao2021learning} to look ahead multiple packing steps to choose the best placement.
However, as the number of possible placements in each packing step is still very large, analyzing multiple future steps are thus still not sufficiently efficient.

Aiming at high efficiency in robotic bin packing, some packing heuristics are proposed to quickly determine the best object placement location by evaluating an objective function.
These approaches have long been a favorite.
For example, \jiahui{the} deepest bottom-left~\cite{baker1980orthogonal,karabulut2004hybrid} suggests placements closer to the deepest bottom-left corner;
heightmap minimization~\cite{wang2019stable} suggests placements with smaller heightmap increment;
and maximum touching area~\cite{wang2010two} finds placements with more contact area with the already-packed objects.
Though some of the heuristics can deal with non-regular objects, existing packing heuristics are insufficient to account for the object compactness in the container, due to the lack of modeling the object geometry and considering near but non-contacting (adjacent) objects.

In this work, we propose SDF-Pack using the truncated signed distance field to model the container's geometry and formulate the SDF-minimization heuristic to encourage nearby objects to be packed more closely, even though they are not directly contacting one another in the container.

%%%%%%%%%%%%%%%%%%%%%%%%%%%%%%%%%%%%%%%%%%%%%%%%%%%%%%%%%%%%%%%%%%%%%%%%%%%%%%%%

\section{METHOD}

\subsection{Problem Setup}
\label{sec:setup}
Given a sequence of \jhs{objects, which can be regular or irregular} our goal is to iteratively suggest a placement location and orientation for each object, such that the packing solution \jhs{can be} more compact, and more objects can be successfully packed into the container.
A good packing solution should be both feasible and compact, and a feasible bin packing solution should satisfy the following conditions:
\begin{itemize}
\item \textit{Containment.}
Objects must be fully inside the container.
\item \textit{Collision-free.}
When putting an object into the container, the robot arm and the object must not collide with any previously-packed object and the container itself.
\item \textit{Stability.}
Each object should remain physically stable after it is placed into the container.
\end{itemize}
Furthermore, a compact packing solution should \jhs{pack} the objects as tightly as possible for \jhs{higher} space utilization.

\begin{figure*}
\begin{center}
  \includegraphics[height=0.36 \linewidth, width=1.0 \linewidth]{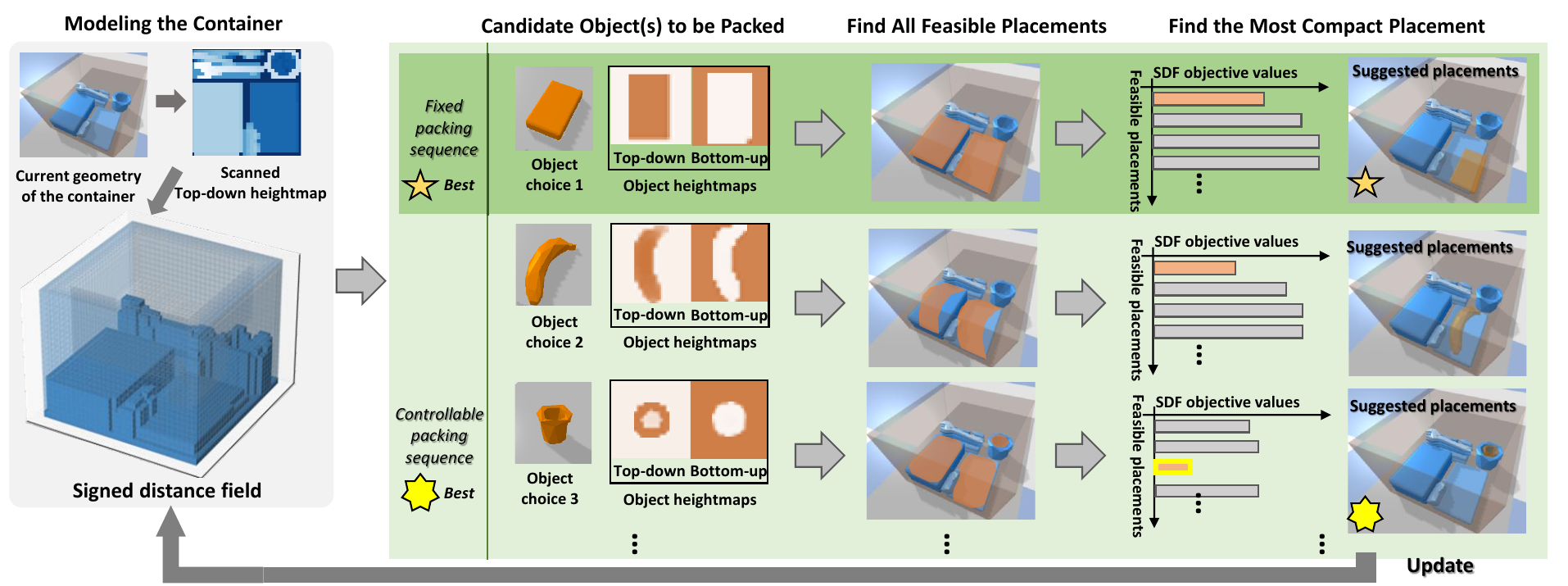}
\end{center}
\vspace*{-4mm}
\caption{Illustrating a packing step in our SDF-Pack.
First, we acquire the top-down heightmap of the container and construct a truncated SDF to model its geometry. 
We represent the candidate objects by their top-down and bottom-up heightmaps, and we find all their feasible placements.
Then, we use our SDF-minimization heuristic to find the most compact placement location, which associates with small nonnegative SDF values. 
For the case of fixed packing order, we need to only consider one object in each step; if the packing order is controllable, we compare the heuristic values of different objects and recommend the best object to achieve a more compact packing.
At last, we go back to update the SDF.
Best viewed in color.
}
\vspace*{-4mm}
\label{fig:framework}
\end{figure*}

\subsection{Our packing framework}
\label{sec:framework}
Figure~\ref{fig:framework} illustrates a single packing step in our framework.
In each step, we scan the current container with the packed objects to \jhs{first} obtain a top-down heightmap, and then construct a signed distance field to represent its geometry.
We leverage the top-down and bottom-up heightmaps to represent each candidate object to be packed and find all feasible placements for each object.
Then, we can make use of the SDF to evaluate all feasible placements and select the one with the lowest SDF objective value to be executed.
Further, when the packing sequence is controllable, we compare the SDF heuristic values of different candidate objects to suggest the best object to be packed to achieve better packing compactness.
For clarity, we present our framework in this section for the case of a controllable packing sequence, which is more general.

In the following, we introduce how we model the container (Section~\ref{sec:bin}) with SDF and how to find feasible placements (Section~\ref{sec:feasible}); after that, we introduce our SDF-minimization heuristic (Section \ref{sec:tsdf}) and additional strategies for controllable packing sequence (Section~\ref{sec:reorder}) and accelerating the computation (Section~\ref{sec:speed}).

\subsection{Modeling the container}
\label{sec:bin}
We follow~\cite{wang2019stable, wang2010two} to scan the top-down heightmap of the container.
The heightmap depicts the 3D volume occupied by previously-packed objects in the container.
We regard the volumes \jiahui{under} the packed objects as occupied, since they are not reachable by the robot arm. Then, from the top-down heightmap, we construct the 3D signed distance field of the container (Figure~\ref{fig:framework} left).
In real-world packing, objects may shift or roll after \jiahui{being put} into the container. So, we re-scan the container after each packing step.

\jiahui{Signed-distance field (SDF)~\cite{gibson1998using,fedkiw2002level} is an implicit representation of the object geometry and is often employed in 3D reconstruction. 
The efficiency of modeling distance field between objects for object manipulation has been noticed recently~\cite{huang2023nift}.}
\jiahui{While in our case, the SDF} denotes the shortest distance from a 3D location in the container to the nearest surface point \jhs{on} the packed object or the container's \jhs{interior}.
A positive (negative) value indicates \jhs{an unoccupied (occupied) location}.
Intrinsically, the SDF describes how close a 3D location is to the existing objects, so we can readily assess the packing compactness of a new object relative to the existing ones in the container.
\jhs{Further}, when estimating the packing compactness of a candidate object placement, we only need to consider the geometry of \jhs{the} nearby objects. 
So, we adopt a truncated SDF, which clamps large distance values to improve the computation efficiency\jhs{; also, it helps to localize the SDF computation and update.}

Compared with previous packing heuristics, the distinctive advantage of our approach is that it can {\it better optimize the object packing compactness\/}.
Using SDF, we can account for the {\it object proximity\/} and consider objects that are {\it nearby but not necessarily contacting\/}, enabling us to readily assess compactness and find more compact object placements.

\begin{algorithm}
\caption{Finding Feasible Placements\label{alg:search}}
\noindent\textbf{inputs:} the top-down heightmap of the container $H_c$ and the bottom-up heightmaps of the object \jiahui{$\hat{H}_o^r$} at different orientations. Z and H are the heights of the container and the object, respectively.
\begin{algorithmic}[1]
\For{all ($x$, $y$, $r$)}
    \State Find the lowest collision-free placement $z$ by Eq. (\ref{eq:feasible}).
        \If{\jiahui{$z+H<=Z$} and StabilityTest($(x,y,z)$, $H_c$, \jiahui{$\hat{H}_o^r$})}
            \State $\mathcal{T} \gets (x,y,r)$;
        \EndIf
\EndFor
\State \textbf{return} $\mathcal{T}$
\end{algorithmic}
\end{algorithm}

\subsection{Finding feasible placements.} 
\label{sec:feasible}
To pack an object (represented by a pair of top-down and bottom-up heightmaps), we need to enumerate all feasible locations \{($x$, $y$, $z$, $r$)\} of placing it in the container, where $x,y$ are the horizontal object coordinates, $z$ is the height level, and $r$ is the object's orientation on the $xy$-plane.
A placement ($x$, $y$, $z$, $r$) is regarded as feasible, if it fulfills the containment, collision-free, and stability constraints.
\jiahui{To obtain} feasible placements,
%\jiahui{that meet} the , 
we find all possible combinations of \jiahui{$x$, $y$, and $r$}, locate the deepest $z$ to place the object without collisions (Eq.~\eqref{eq:feasible}), \jiahui{and} then perform the stability test (Algorithm~\ref{alg:stable}). 
%
%A placement ($x$, $y$, $z$, $r$) that fulfills all constraints is considered as feasible.
%
%The procedure is shown in Algorithm~\ref{alg:search}.
Algorithm~\ref{alg:search} shows the procedure.

Similar to~\cite{wang2019stable}, we leverage the top-down heightmap of the container and the bottom-up heightmap of the candidate object at the orientation $r$ to find the deepest reachable $z$:
\begin{equation}
z = G(x,y,r) = \max^{W-1}_{i=0}{\max^{D-1}_{j=0}{(H_{c}[x+i,y+j] - \hat{H}_o^r[i,j])}},
\label{eq:feasible}
\end{equation}
where $H_{c}$ is the top-down heightmap of the container;
$\hat{H}_o^r \in \mathbb{R}^{W \times D \times H}$ is the bottom-up heightmap of the input object at orientation $r$; and
\jiahui{$W$, $D$, and $H$ are the width, depth, and height of the object.}
The heightmap of each view is measured towards the opposite plane of the object's bounding box.

\begin{algorithm}
\caption{StabilityTest}\label{alg:stable}
%\noindent\textbf{Input:} Target placement $(x,y,z)$, container's top-down heightmap $H_B$ and  object's bottom-up heightmaps $\hat{H}_O^r$ at in-plane rotation $r$. \\
%\noindent\textbf{Output:} True/False
\begin{algorithmic}[1]
%\Procedure{StableTest}{}
\State $ \{(p_x, p_y, p_z)\} \gets \hat{H}_o^r \cap H_{c[x:x+W, y:y+D, z-1:z+H-1]}$ \newline \Comment{\textit{Compute supporting points for the object.}}

\State $ \mathcal{P} \gets ConvexHull(\{p_x, p_y\}) $ \newline\Comment{\textit{Construct the support polygon}}
\If {$(m_x, m_y) \in Int(\mathcal{P})$}
    \State \textbf{return} True
\Else
    \State \textbf{return} False
\EndIf
\newline\Comment{\textit{If the object's mass center locates in the support polygon, then it is stable.}}
\end{algorithmic}
\end{algorithm}

Then, we measure the stability of the object placed at ($x$, $y$, $z$, $r$) by checking if the object's mass center lies inside its support polygon~\cite{hu2020tap, ramos2016container}. 
The stability test is shown in Algorithm \ref{alg:stable}. 
First, we retrieve the supporting points of the object, i.e., the points that the object contacts the bottom or \jhs{already-packed} objects in the container. 
Then, we project the supporting points and the object's mass center to the $xy$-plane.
If the projected mass center locates in the convex hull of the projected supporting points (i.e., the support polygon~\cite{hu2020tap, ramos2016container}), the placement is regarded as stable.
%
%\jiahui{Although due to the complex inter-object tilted forces, passing the stability test does not guarantee complete object stability in real-world scenarios, the support polygon test aids in eliminating non-stable placements.}
\jiahui{Due to the complex tilted forces between the non-regularly shaped objects, passing the stability test may not always guarantee stable placement in the real world, yet this support polygon test can quickly eliminate major non-stable placements.}
A candidate placement is regarded as feasible only if the object does not exceed the container's top and \jiahui{it passes the stability test.}
Lastly, we choose the optimal placement by finding the feasible placement with the optimal packing objective value.

\subsection{SDF-minimization heuristic}
\label{sec:tsdf}

We formulate the following \textit{SDF-minimization heuristic} consisting of three terms to evaluate the object packing compactness (lower is better) locally around each obtained feasible placement $(x, y, z, r)$ for an object $o$.

\vspace*{-2.5mm}
{
\small
\begin{equation}
\begin{aligned}
F_o(x, y, z, r) = 
&
\frac{\alpha}{V_o} \sum_{w=0}^{W-1}{\sum_{d=0}^{D-1}{\sum_{h=0}^{H-1}{\Phi(x+w,y+d,z+h) \cdot O_o^r(w,d,h)}}} \\
& + \beta (1 - \sqrt[3]{\frac{V_o}{W \cdot D \cdot H}}) + \gamma \cdot z, \\
\end{aligned}
\label{eq:sdfm}
\end{equation}
}

\noindent
where
$\alpha$, $\beta$, and $\gamma$ are \jiahui{the} weights on the three terms, respectively;
$\Phi(\cdot)$ is the truncated SDF;
$W$, $D$, and $H$ are the width, depth, and height of the object at orientation $r$;
$O_o^r(\cdot)$ records the occupancy of the object \jhs{(its value is $1$ for occupied location and $0$ otherwise)};
$V_o$ is the total volume of the object. 
All of $W$, $D$, $H$, $O_o^r(\cdot)$ and $V_o$ can be obtained using the heightmaps of object $o$.

The first term computes the average TSDF value of the volumes occupied by the object.
By minimizing this term, we can find a candidate location that places the object more compactly around the already-packed objects and the container walls.
The second term further encourages axis-aligned object placements for regularly-shaped objects.
Here, we measure the volume ratio of the object to its axis-aligned bounding box, so minimizing this term encourages a more axis-aligned object orientation.
Lastly, to further maximize the space utilization for accommodating more objects in the future, we encourage the placement to be as deep as possible by minimizing the $z$ value of the placement.

\subsection{Extension to controllable packing sequence.}
\label{sec:reorder}
The SDF-minimization heuristic $F_o$ in Eq.~\eqref{eq:sdfm} can be extended to handle \jiahui{a} controllable packing sequence, in which we are allowed to choose \jhs{which} candidate object to be packed first to further improve the packing compactness.
One simple approach is to evaluate $F_o$ for all candidate objects, and then select the one and its associated placement with the lowest $F_o$ value to execute.
However, this approach may not be optimal, as picking small objects too early may break the container space into fragments that large objects may not easily fit into, thereby lowering the overall space utilization.
Hence, we account for objects of varying sizes by adding a size-balancing term to try to pick larger objects earlier, and pick small objects unless they can well fit holes and gaps around the existing objects in the container:

\vspace*{-2.5mm}
{
% \small
% \begin{small}
\begin{equation}
\begin{aligned}
\hat{F_o}(x, y, z, r) 
=
F_o(x, y, z, r) + \delta \cdot (1 - \sqrt[3]{\frac{V_o}{X \cdot Y \cdot Z}}), \\
\end{aligned}
\label{eq:reorder}
\end{equation}
}

\noindent
where $\delta$ is a weight; and $X$, $Y$, and $Z$ are the width, depth, and height of the container.
Note that minimizing the last term essentially maximizes the volume ratio of the selected object to the container, preferring larger objects as a result. 
Our method can flexibly re-plan the packing order on the fly.
When the number of unpacked objects is extremely large, we simulate a buffer as in many real-world packing scenarios and re-plan the packing sequence of the first K objects.

\subsection{Improving computation efficiency}
\label{sec:speed}

\noindent\textbf{GPU computation for truncated SDF field construction.}
Constructing the truncated SDF (TSDF) field is to determine the distance from each location in the container to the closest occupied location within the truncated distance.
Such a computation process can be parallelized by sliding a 3D kernel in the container and accumulating the shortest distance.
To this end, we develop a GPU implementation in PyTorch~\cite{paszke2019pytorch} to speed up the process.
Note that we only \jhs{perform} GPU computation in the TSDF construction, and we still use CPU sequential computation in a single thread for a \jhs{fair} comparison with other methods.

Doing so helps to \jhs{avoid} repetitive computation and speed up the feasibility test (Section~\ref{sec:feasible}) and heuristic computation (Sections~\ref{sec:tsdf} and \ref{sec:reorder}).
After each packing step, since the effect of the new placement is local, we only update the feasible placements and heuristic values within the truncated distance around a 2D bounding box, in which the heightmap values have changed after the last packing step.

%%%%%%%%%%%%%%%%%%%%%%%%%%%%%%%%%%%%%%%%%%%%%%%%%%%%%%%%%%%%%%%%%%%%%%%%%%%%%%%%
\section{EXPERIMENT}
\subsection{Dataset}
\begin{figure}[t]
\begin{center}
  \includegraphics[scale=0.9]{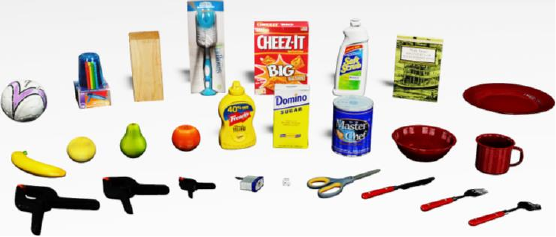}
\end{center}
\vspace{-4mm}
\caption{Examples of some real-world irregular objects in our dataset.} 
\vspace{-2mm}
\label{fig:dataset}
\end{figure}

To test the packing performance of our method, we build a dataset that contains 96 types of real-world objects: 71 from the YCB dataset~\cite{ycb} and 25 from the Rutgers APC RGB-D dataset~\cite{apc}.
Figure~\ref{fig:dataset} shows some of these objects.
Considering the picking dynamics of the robot arm with a suction cup, we re-orientate some objects to ensure that all objects \jiahui{remain stable on a horizontal plane and} are graspable in the vertical direction.
Due to the robot arm movement, only horizontal in-plane rotations are allowed on the objects during the packing procedure.
We pre-process the 3D models of all objects using~\cite{huang2018robust}. 
Each processed mesh has at most 100 vertices and 150 faces.
Further, we perform convex decomposition on them using the V-HACD \cite{mamou2016volumetric} algorithm to enable a more realistic collision effect in our experiments with physical simulation.

\subsection{Implementation details}
\label{sec:imp_details}
\noindent\textbf{Packing environment setup.} 
Following~\cite{wang2019stable}, we consider a container of size $32 cm \times 32 cm \times 30 cm$ \jiahui{with} resolutions of $0.01 m$ \jhs{in} $x$ and $y$ dimensions and $0.002 m$ in \jiahui{the} $z$ dimension to discretize the scanned heightmaps of the container and the objects when constructing the truncated SDF field.
\jiahui{Since many objects from the dataset are centrosymmetric (e.g., the mustard bottle, bowl, etc.), w}e search for the object's \edward{possible} $xy$-plane orientations \edward{$(r)$} for every $\pi/4$ within $[0,\pi)$ \jiahui{for efficiency}.
The truncate distance for the SDF field is set as five \edward{units}.
In our implementation, the hyper-parameters in Eqs.~\eqref{eq:sdfm} and~\eqref{eq:reorder} are set as $\alpha=2.5$, $\beta=10.0$, $\gamma=1.0$, and $\delta=80.0$.
\jhs{Our GPU computation for the TSDF construction is performed on a single NVIDIA TITAN Xp.}
%{\it We will release the code of our method after publication.}

\vspace{+8pt}
\noindent\textbf{Physical simulation.} 
All our packing experiments are performed in the PyBullet~\cite{coumans2021} physical simulator.
We set the gravity as $-9.8 m/(s^2)$ and the mass value of each object is set randomly around $(0,1)$ per ${cm}^3$.
After packing an object, we wait \jhs{for} $0.25$ seconds until it is stable. Then, we scan the container's top-down heightmap in the next packing step.

\vspace{+8pt}
\noindent\textbf{Packing cases.}
We perform experiments on 1,000 object sequences randomly generated using $80$ everyday objects, which is four times compared with \cite{wang2019stable}, for evaluating different methods towards the packing limit of the container.
In the experimental setting of \cite{wang2019stable}, they run through the objects in the sequence one by one and skip an object if it does not have a feasible placement.
In comparison, since we cannot skip too many unpacked objects in the real-world packing scenario,
we stop a packing procedure for the current container early if in total $K$ objects cannot find a feasible placement. 
Otherwise, the packing procedure finishes when all objects in the sequence have been evaluated.

\vspace{+8pt}
\noindent\textbf{Fixed and controllable packing sequences.}
We perform experiments in two packing settings with (i) \edward{a} \textbf{fixed} packing sequence, in which we must follow the order of the randomly generated sequences to pack the objects and (ii) \edward{a} \textbf{controllable} packing sequence, in which we can partially re-arrange the packing order. 
\edward{In the latter setting,} \jiahui{as in some packing scenarios~\cite{zhao2021online},} we set a buffer of size $K$ to store the first $K$ arriving objects, select an object from the buffer one at a time\jiahui{, and then fill the buffer with the next arriving object.}
\jiahui{In other words, t}he packing order of the buffered objects can be re-arranged using some rules (e.g., volume decreasing order~\cite{wang2019stable}) or using some objectives (e.g., Eq.~\ref{eq:reorder}) (see Section~\ref{sec:experiment_control}).
We set $K=5$ in our experiments, and the former setting is equivalent to setting a buffer size of one.

\vspace{+8pt}
\noindent\textbf{Evaluation metrics.}
We run all methods in the same packing environment, physical simulation, and packing sequences.
We evaluate various packing methods using (i) \textit{packed volume \jiahui{(or packed vol.)}}, \edward{\ie,} the total volume of the successfully-packed objects measured in $cm^3$; (ii) \textit{compactness}, \edward{\ie,} the average ratio of the total volume occupied by all the packed objects over the \edward{volume} of the overall bounding box
containing all \jhs{the} packed objects; and (iii) \textit{packed object number}, \edward{\ie,} \jhs{the} number of objects successfully packed in the container. \jiahui{In addition, to better indicate the performance of our method and other existing packing heuristics, we also show the increment of the packed volume (vol. inc.) for each method towards the deepest bottom-left (DBL)~\cite{baker1980orthogonal,karabulut2004hybrid}.}

\begin{table}[t] 
  \centering
  \begin{minipage}{\linewidth}
  \caption{Packing results on 1,000 packing cases of different packing heuristics with fixed packing sequences}
  \label{tab:online}
  \resizebox{\linewidth}{!}{
  \begin{tabular}{@{\hspace*{0mm}}l@{\hspace*{1mm}}@{\hspace*{0mm}}|@{\hspace*{2mm}}c@{\hspace*{2mm}}c@{\hspace*{2mm}}c@{\hspace*{2mm}}c@{\hspace*{0mm}}}
    \hline
    & \jiahui{\textbf{Packed vol.}} & \textbf{Vol. inc.} & \textbf{Compactness} & \textbf{Obj. num.} \\% &  \textbf{Time per obj.} \\
    \hline
    \textbf{Random} & 8786.31 & -37.1\% & 0.35 & 23.05\\%  & 0.03\\
    \textbf{FF~\cite{falkenauer1996hybrid}} & 12435.91 & -11.0\% & 0.43 & 32.34\\%  & 0.15\\    
    \hline
    \textbf{DBL~\cite{baker1980orthogonal,karabulut2004hybrid}} & 13971.31 & 0.0\% & 0.47 & 34.85\\%  & 0.33\\
    \textbf{MTA~\cite{wang2010two}} & 14002.52 & 0.2\% & 0.47 & 35.30\\%  & 0.36\\
    \textbf{HM~\cite{wang2019stable}} & 14482.61 & 3.7\% & 0.49 & 36.34\\%  & 0.35\\
    \textbf{SDF-Pack (Ours)} & \textbf{15054.56} & \textbf{7.8\%} & \textbf{0.51} & \textbf{37.38}\\%  & 0.54\\
    \hline
    \end{tabular}}
    \vspace{-10pt}
\end{minipage}
\end{table}

\subsection{Comparison on fixed packing sequence}
We first compare our method for the scenario of fixed packing sequence on 1,000 packing cases against the existing heuristics: deepest bottom-left (DBL)~\cite{baker1980orthogonal,karabulut2004hybrid,wang2010two}, maximum-touching-area (MTA)~\cite{wang2010two}, and heightmap-minimization (HM)~\cite{wang2019stable}.
All compared methods share the same inputs (i.e., a top-down heightmap for the container, a top-down and a bottom-up heightmaps for each object) and the same feasible placement search procedure. The only difference is that they use different objective functions to find the best placement for each object.
We also compare our method with random placement (Random), which \jiahui{randomly selects a feasible placement,} and the first-fit placement (FF)~\cite{falkenauer1996hybrid}, which selects the first feasible placement.

Table~\ref{tab:online} reports the experimental results, showing that our method achieves a better packing performance than the compared methods. 
On average, our method achieves a gain of $7.8\%$ on the packed volume compared with \jiahui{DBL}, exceeding all compared heuristics. 
Also, it improves the packing compactness, \jiahui{packing} one to three more objects compared with the existing heuristics.
SDF-Pack's computation time is only $0.54$s per object,  similar to other methods ($0.36$s for MTA, $0.35$s for HM, and $0.33$s for DBL), which is neglectable compared with the robot arm's movement time.

\begin{table}[t] 
  \centering
  \begin{minipage}{\linewidth}
  \caption{Packing results on 1,000 packing cases for different packing heuristics with controllable packing sequences, following bounding-box volume decreasing order.}
  \vspace{-2mm}
  \label{tab:voldec}
  \resizebox{\linewidth}{!}{
  \begin{tabular}{@{\hspace*{0mm}}l@{\hspace*{1mm}}@{\hspace*{0mm}}|@{\hspace*{2mm}}c@{\hspace*{2mm}}c@{\hspace*{2mm}}c@{\hspace*{2mm}}c@{\hspace*{0mm}}}
    \hline
  & \jiahui{\textbf{Packed vol.}} & \textbf{Vol. inc.} & \textbf{Compactness} & \textbf{Obj. num.}\\%  & \textbf{Time per obj.} \\
    \hline
    \textbf{Random} & 10954.60 & -24.09\% & 0.40 & 27.59\\%  & 0.03\\
    \textbf{FF~\cite{falkenauer1996hybrid}} & 13309.71 & -7.78\% & 0.45 & 33.81\\%  & 0.22\\
    \hline
    \textbf{DBL~\cite{baker1980orthogonal,karabulut2004hybrid}} & 14432.86 & 0.0\% & 0.49 & 35.53\\%  & 0.50\\
   \textbf{MTA~\cite{wang2010two}} &14642.87 & 1.5\% & 0.49 & 36.41\\%  & 0.55\\
 \textbf{HM~\cite{wang2019stable}} & 14878.36 & 3.1\% & 0.50 & 36.91\\%  & 0.57\\
   \textbf{SDF-Pack (Ours)\footnote[1]{\scriptsize{\jiahui{For a fair comparison, instead of using Eq.~\ref{eq:reorder}, we re-arrange the packing sequence in the buffer by the bounding-box volume decrease order as in \cite{wang2019stable}, and pack the first feasible buffered object using Eq.~\ref{eq:sdfm} in each packing step.}}}}  & \textbf{15395.61}  & \textbf{6.7\%} & \textbf{0.52}  & \textbf{37.90}\\%   & 0.70\\
    \hline
    \end{tabular}}
    \vspace{-10pt}
\end{minipage}
\end{table}

\begin{table}[t] 
  \centering
  \begin{minipage}{\linewidth}
  \caption{Packing results on 1,000 packing cases for different packing heuristics with controllable packing sequences, using packing order planned by each heuristic.}
  \vspace{-2mm}
  \label{tab:offline}
  \resizebox{\linewidth}{!}{
  \begin{tabular}{@{\hspace*{0mm}}l@{\hspace*{1mm}}@{\hspace*{0mm}}|@{\hspace*{2mm}}c@{\hspace*{2mm}}c@{\hspace*{2mm}}c@{\hspace*{2mm}}c@{\hspace*{0mm}}}
    \hline
  & \jiahui{\textbf{Packed vol.}} & \textbf{Vol. inc.} & \textbf{Compactness} & \textbf{Obj. num.}\\%  & \textbf{Time per obj.} \\
  \hline
\textbf{DBL~\cite{baker1980orthogonal,karabulut2004hybrid}\footnote[2]{\label{foot:balance}\scriptsize{\jiahui{For a fair comparison, we incorporate the size-balancing term from Eq.~\ref{eq:reorder} to each compared heuristic, and in each packing step, we select the buffered object and its corresponding placement with the optimal heuristic value to pack. 
}}}}  & 14914.44  & 0.0\% & 0.50  & 35.60\\%   & 1.35\\
   \textbf{MTA~\cite{wang2010two}\textsuperscript{\ref{foot:balance}}}  & 12463.57  & -16.4\% & 0.44  & 27.81\\%   & 1.73\\

   \textbf{HM~\cite{wang2019stable}\textsuperscript{\ref{foot:balance}}}  & 13910.88  & -6.7\% & 0.49  & 37.86\\%   & 1.21\\
   \textbf{SDF-Pack (Ours)}  & \textbf{16363.34}  & \textbf{9.7\%}  & \textbf{0.55}  & \textbf{39.36} \\% & 1.45\\
    \hline
    \end{tabular}}
    \vspace{-10pt}
\end{minipage}
\vspace{-10pt}
\end{table}

\begin{table}[t] \scriptsize
  \centering
  \caption{Comparing our method with a Genetic-algorithm-based method on 200 random
  packing cases with controllable packing sequences.}
  \vspace{-2mm}
  \label{tab:genetic}
  \resizebox{0.8\linewidth}{!}{
  \begin{tabular}{l|cc}
    \hline
    & \textbf{Genetic-DBL} & \textbf{Ours}\\
   \hline
    \textbf{\scriptsize{Packed volume}}& 15787.68 & 16357.28\\
    \textbf{\scriptsize{Compactness}}& 0.52  & 0.56\\
    \textbf{\scriptsize{Packed object number}}& 45.70 & 39.44\\
    \textbf{\scriptsize{Time per object (seconds)}}& 22.84 & 1.47\\
    \hline
    \end{tabular}}
  \vspace{-2mm}
\end{table}

\subsection{Comparison on controllable packing sequence}
\label{sec:experiment_control}
\noindent\textbf{Bounding-box volume decreasing order.} Second, we compare our SDF-Pack with the same set of methods on 1,000 random packing cases in the case of controllable packing sequences (see Section \ref{sec:imp_details}).
Given the $K$ objects in the buffer, we first follow \cite{wang2019stable} to sort the objects in descending order of \jiahui{their} bounding-box volume, \jiahui{and} then pack the first feasible object in each packing step.
After that, we fill the buffer with the next object in the packing sequence and sort the buffered objects again.

As shown in Table~\ref{tab:voldec}, our method outperforms others in terms of packed volume, achieving $6.7\%$ larger than DBL.
Further, our method achieves the best packing compactness and can pack \jhs{one to ten} more objects into the container when compared to different approaches. 
The results demonstrate the effectiveness of our approach in finding a better placement location, even when only using a fixed rule to select the next object in the buffer.

\vspace{+8pt}
\noindent\textbf{Re-planned packing order using Eq.\eqref{eq:reorder}.} 
Then, we evaluate our SDF-minimization heuristic that uses Eq.~\eqref{eq:reorder} to select the optimal buffered object instead of using the above rule.
We compare our complete heuristic with DBL~\cite{baker1980orthogonal,karabulut2004hybrid}, MTA~\cite{wang2010two}, and HM~\cite{wang2019stable}.
For a fair comparison, we incorporate the size-balancing term into each compared heuristic. 
Specifically, for the heuristics that prioritize the lowest objective value (i.e., DBL and HM), we add the size-balancing term as in our method; for MTA, which prioritizes the highest objective value, we subtract the term instead.

Table~\ref{tab:offline} shows that our complete heuristic further improves the packing volume by \jiahui{6\% in the setting that allows re-planning the packing order (see the last rows of Tables~\ref{tab:voldec} and~\ref{tab:offline})}. %\jiahui{(16363.34 v.s. 15395.61)}. 
Also, our SDF-minimization heuristic outperforms the compared methods by at least $9.7\%$ in terms of packed volume, over $5\%$ in terms of compactness, and over $1.5$ in terms of packed object number, giving clear evidence of the effectiveness of our complete SDF-minimization heuristic.
MTA and HM perform worse than DBL because MTA may leave a large object unpacked if it has a very limited exact contact area (e.g., balls), and HM inherently tends to place small objects first because they have a small heightmap increment.
On average, our method suggests the \edward{best} object and \edward{the} corresponding placement in just $1.45$s, which is slightly faster than MTA ($1.73$s) and comparable with DBL ($1.35$s) and HM ($1.21$s).
The efficiency advantage of our method is that we pre-compute the SDF field at the start of each packing step (see Section~\ref{sec:speed}) and only need to sum up the SDF values for each object and each placement. 

\vspace{+8pt}
\noindent\textbf{Compared with a genetic-algorithm-based method.}
Also, we compare our method on 200 random packing cases with~\cite{kang2012hybrid}, which uses a genetic algorithm to re-plan the whole packing sequence. 
In short, the method adopts DBL to suggest each placement location and uses the genetic algorithm to find the order.
Due to the long computation time required by the genetic method, we only perform 10 iterations for the genetic method.
As Table~\ref{tab:genetic} shows, to make genetic-DBL achieve similar performance to our method, it takes over 15x of computation time. 
This can be an issue when applied to real usage, since the robot has to wait for the computation.
In comparison, our SDF-minimization heuristic can better re-plan the packing order of irregular objects \textbf{almost on the fly} with a lower computation cost, since it does not require trying multiple packing orders to get the overall packing.

\begin{figure*}
\centering
  \includegraphics[ width=2\columnwidth]{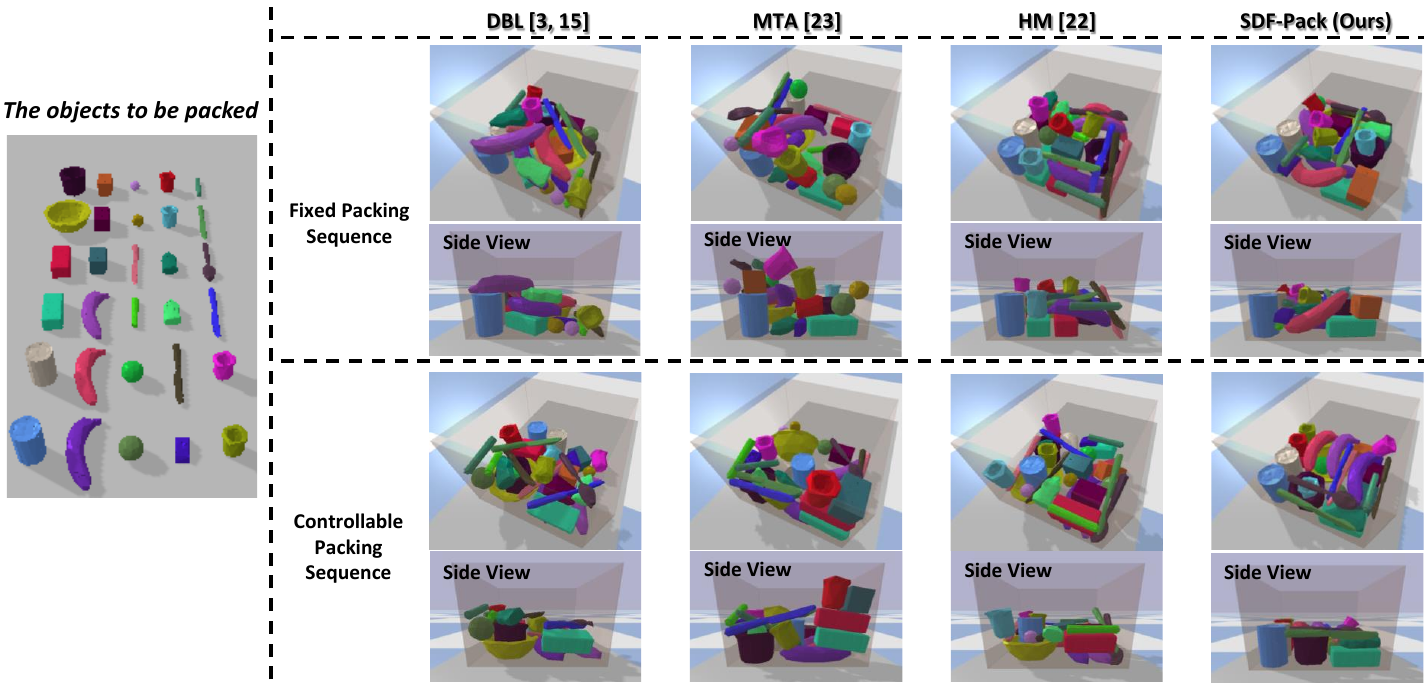}
  \vspace{-8pt} 
\caption{Visual comparison of packing results produced by different methods.
    Using the same sequence of objects, the packing results produced by our SDF-Pack are consistently more compact for both the fixed and the controllable packing sequences.}
\vspace{-8pt} 
\label{fig:qualitative}
\end{figure*}

\subsection{Ablation Study}
We perform an ablation study on 200 random packing cases to explore the effect of each term in our SDF-minimization heuristic.
We perform experiments in both cases of both fixed and controllable packing sequences, using the heuristic value to suggest the placement and the object to be packed.
From Table~\ref{tab:ablation}, we can observe a drop in the packed volume after removing the TSDF term in both fixed (Rows 2 v.s. 1) and controllable (Rows 4 v.s. 3) packing sequences.
Removing both the TSDF and Regular terms reduces the packed volume on average by $4\%$ (Rows 3 v.s. 1) in the setting of fixed packing sequence.
A performance drop is also seen when removing the size-balancing term in Eq.~\eqref{eq:reorder} (Rows 6 v.s. 4, about -7.7\%).
These results show that all proposed terms in the heuristic contribute to the final performance of our packing framework in different settings.

\begin{table}[t]
  \centering
  \caption{The ablation study on 200 \jhs{random packing cases} with fixed and controllable packing sequences.}
  \label{tab:ablation}
  \resizebox{\linewidth}{!}{
  \begin{tabular}{@{\hspace*{0mm}}l@{\hspace*{0mm}}@{\hspace*{2mm}}|@{\hspace*{2mm}}c@{\hspace*{2mm}}c@{\hspace*{2mm}}c@{\hspace*{0mm}}}
  \hline
  & \jiahui{\textbf{Packed vol.}} & \textbf{Compactness} & \textbf{Obj. num.}\\
  \hline
  \textbf{[Fixed] Full model} & 15045.44 & 0.51 & 37.52 \\
  \textbf{[Fixed] w/o TSDF (Eq.\ref{eq:sdfm})}  & 14624.83 & 0.50 & 36.68\\
  \textbf{[Fixed] w/o TSDF \& Regular (Eq.\ref{eq:sdfm})}  & 14440.32  & 0.49  & 36.07\\
  \hline
  \textbf{[Controllable] Full model} & 16357.28 & 0.56 & 39.44 \\
  \textbf{[Controllable] w/o TSDF (Eq.\ref{eq:reorder})} & 15878.73 & 0.54 & 38.55\\
  \textbf{[Controllable] w/o size-balancing (Eq.\ref{eq:reorder})} & 15092.35 & 0.52 & 39.91\\
  \hline
  \end{tabular}}
  \vspace{-10pt}
\end{table}

\subsection{Qualitative analysis}
For ease of visualization, we show a visual comparison with a reduced object set (30 objects).
As shown in Figure~\ref{fig:qualitative}, our method achieves a more compact packing consistently for both controllable and fixed packing sequences.
In comparison, other approaches either cannot effectively utilize the empty space near the upper-right corner of the container, or tend to place objects right next to the container's \jhs{interior} to maximize the touching area leaving the central area not well utilized.
%
%\jiahui{More visual comparisons are shown in the supplementary video.}
More visual comparison results can be found in the supplementary material.

\begin{figure*}
\begin{center}
  \includegraphics[width=1.9\columnwidth]{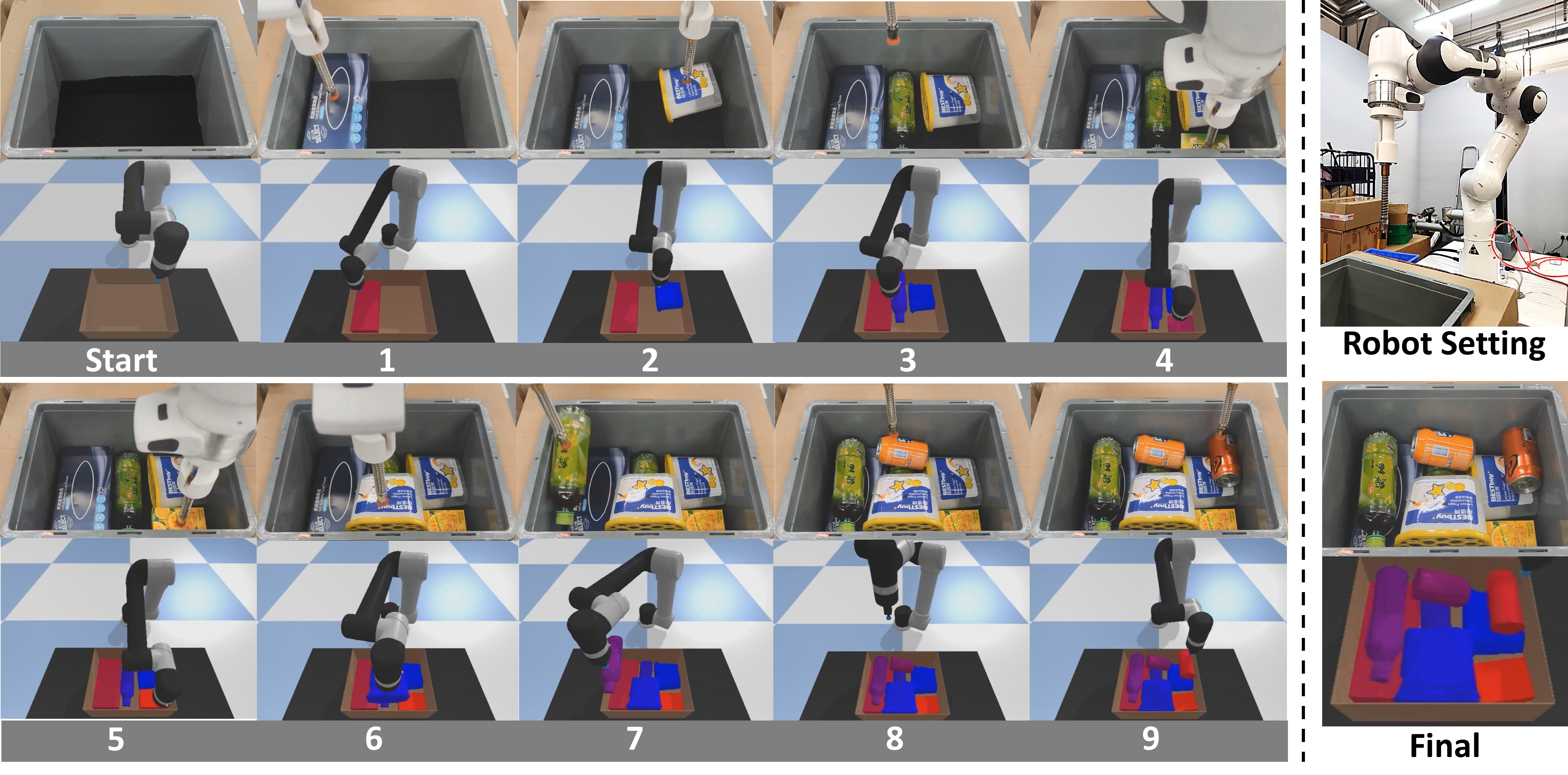}
\end{center}
\vspace{-5mm}
\caption{Visualizing a real-world packing procedure produced by our method.}
\label{fig:real_demo}
\vspace{-6mm}
\end{figure*}

\subsection{Robotic demos}
Further, we set up a real-world robotic packing scene using the Franka Emika Robot with a suction cup and apply our method to pack everyday supermarket products in this environment.
Besides, we build a virtual packing scene based on the Ravens~\cite{zeng2020transporter} for a virtual-to-real comparison. 
The procedure of using our heuristic to pack nine supermarket products \jhs{into a container of size $45cm \times 32cm \times 20cm$} is shown in Figure~\ref{fig:real_demo}. 
From the figure, we can see that our heuristic helps to better utilize the container space. 
It first packs the tissue box compactly towards the container's \jhs{interior} (since it is the largest with the lowest SDF value), making full use of the bottom-left space in the container.
Then, it compactly fills the container's bottom layer by putting the dehumidifier, the boxed lemon tea, etc. 
Please refer to the supplementary video for more results.

\section*{Conclusion}
We presented SDF-Pack, a new approach to enhance robotic bin packing with the truncated signed distance field to model the container's geometric condition and the SDF-minimization heuristic to effectively assess the spatial compactness and find compact object placements.
Experimental results manifest that SDF-Pack can consistently achieve the highest packing performance compared with all existing heuristics for both packing scenarios with the fixed and the controllable packing sequences.

In the future, we plan to explore the followings: first, how to improve the robustness of our packing computation when given heightmaps with noise and how to generalize to non-rigid deformable objects; and
\jiahui{second, how to integrate our SDF-based objective with reinforcement learning, to generate a better packing sequence by adjusting the object packing order and more optimally utilize the container space.}
%
%Further, we plan to explore repetitive local geometric patterns when packing objects in the container, particularly we would like to leverage a data-driven approach and integrate our novel SDF-based objective into the training process.

%%%%%%%%%%%%%%%%%%%%%%%%%%%%%%%%%%%%%%%%%%%%%%%%%%%%%%%%%%%%%%%%%%%%%%%%%%%%%%%%

{\small
\normalem
\bibliographystyle{IEEEtranS}
\bibliography{egbib}
}

\end{document}